 \documentclass[tablecaption=bottom,wcp,usenames,dvipsnames]{jmlr} % W&CP article

\usepackage{preamble}
\usepackage{maths_preamble}

\usepackage[load-configurations=version-1]{siunitx} % newer version
\theorembodyfont{\upshape}
\theoremheaderfont{\scshape}
\theorempostheader{:}
\theoremsep{\newline}

\jmlrproceedings{AABI 2023}{5th Symposium on Advances in Approximate Bayesian Inference, 2023}

\title[Amortised Inference in Neural Networks]{Amortised Inference in Neural Networks for Small-Scale Probabilistic Meta-Learning}

 \author{\Name{Matthew Ashman}$^*$ \Email{mca39@cam.ac.uk}\\
  \Name{Tommy Rochussen}$^*$ \Email{tnr22@cam.ac.uk}\\
  \Name{Adrian Weller} \Email{aw665@cam.ac.uk}\\
  \addr University of Cambridge\\
  \addr * Equal contributions}

\begin{document}

\maketitle

% \def\thefootnote{*}\footnotetext{These authors contributed equally to this work}

% \begin{abstract}
% This is the abstract for this article. It is optional for AABI submissions.
% \end{abstract}

% Keywords may be removed
%\begin{keywords}
%List of keywords
%\end{keywords}

\section{Introduction}
\label{sec:intro}

In many machine learning applications, well-calibrated posterior predictive distributions are required for a number of closely-related datasets. Given similarity between datasets, it is natural to wish to develop meta-learning algorithms that utilise other datasets to reduce the computational complexity and / or improve predictive performance when deploying models on newly-seen datasets at test time. There have been a number of significant recent developments in meta-learning for predictive distributions, most notably that of the neural process (NP) family \citep{garnelo2018conditional,garnelo2018neural,foong2020meta,gordon2018meta,gordon2019convolutional}. Despite the utility of these methods on large-scale meta-datasets, they perform poorly in settings where the number of datasets and the total number of datapoints is small. We argue that this is a result of the large number of shared model parameters
%---needed due to inefficient model structure---
 overfitting to the meta-dataset. A natural solution is to 
 % improve model structure and 
  remove these shared model parameters, and instead train a meta-model to learn to approximate fully Bayesian inference over task-specific model parameters.

Recently, \cite{ober2021global} developed a variational approximation for Bayesian neural networks (BNNs) based on using a set of inducing inputs to construct a series of conditional distributions that accurately approximate the conditionals of the true posterior distribution. Notably, the variational distribution consists of the prior multiplied by a set of approximate likelihoods for each inducing input. Our key insight is that these inducing inputs can be replaced by the actual data, such that the variational distribution consists of a set of approximate likelihoods for each datapoint. This structure lends itself to amortised inference, in which the parameters of each approximate likelihood are obtained by passing each datapoint through a meta-model known as the inference network. By training this inference network across related datasets, we can meta-learn Bayesian inference over task-specific BNNs, addressing the challenge above.

\section{Related Work}
\paragraph{Neural processes}
Our work is most similar to the NP family \citep{garnelo2018conditional, garnelo2018neural}, which seeks to meta-learn predictive distributions either through maximisation of the posterior predictive likelihood or variational inference (VI) \citep{foong2020meta}. Similar to our method, NPs utilise an encoder to create embeddings for each datapoint. These embeddings are then aggregated to form a distribution over a latent variable which is then sampled and passed, together with a test datapoint, through a decoder. \cite{volpp2020bayesian} propose the use of Bayesian aggregation, in which embeddings of individual datapoints take the form of approximate likelihoods which are multiplied together with the prior to form an approximate posterior distribution over the latent variable. Whilst these methods differ to ours in their use of shared model parameters, our method can be reinterpreted as a member of the NP family in which the latent variables are the parameters of the decoder. Through this perspective we can train our model in an identical way to NPs.

\paragraph{Meta-learning neural networks}
Meta-learning for neural networks had received a significant amount of attention from the research community. Notable examples include MAML \cite{finn2017model} and its extensions \citep{yoon2018bayesian, antoniou2018train}, which seek good parameter initialisation, and those which explicitly condition on the dataset to obtain task specific parameters \citep{requeima2019fast, gordon2018meta}. Whilst conceptually similar to our approach, these methods differ in their use of shared model parameters---the task specific parameters amount to a small subset of the overall model parameters. In addition to requiring a large meta-dataset, this limits these methods to meta-datasets in which the individual datasets are very similar. By contrast, our approach does not use any shared model parameters but rather meta-learns inference in a BNN. We discuss the relationship between our method and NPs in \Cref{app:neural_processes}, demonstrating that they conceptually differ only in a change of objective functions.

\section{Background}
In this section we review the GI-BNN of \cite{ober2021global} and NP family \citep{garnelo2018neural}. Throughout, let $\*\Xi = \{\mcD\}$ denote a meta-dataset of $|\Xi|$ datasets, and $\mcD = \{\bfX, \bfy\}$ denote a dataset consisting of inputs $\bfX \in \R^{N\times D_0}$ and outputs $\bfy \in \R^{N \times P}$.

\subsection{Global Inducing Point Variational Posteriors for BNNs}
\label{subsec:gibnn}
Let $\bfW = \{\bfW^{\ell}\}_{l=1}^L$ denote the weights of an $L$-layer neural network, such that $\bfW^{\ell} \in \R^{D^{\ell - 1} \times D^{\ell}}$ where $D^{\ell}$ denotes the dimensions of the $\ell$-th hidden layer, and $\psi(\cdot)$ denote the element-wise activation function acting between layers. \cite{ober2021global} introduce the global inducing point variational approximation for BNNs, in which the variational approximation to the posterior $p(\bfW | \mcD)$ is defined recursively as $q_{\phi}(\bfW) = \prod_{\ell = 1}^L q_{\phi}(\bfW^{\ell} | \{\bfW^{\ell'}\}_{\ell' = 1}^{\ell - 1}, \bfU^0)$, where
\begin{equation}
    q_{\phi}\left(\bfW^{\ell} | \{\bfW^{\ell'}\}_{\ell' = 1}^{\ell - 1}, \bfU_0\right) \propto \prod_{d = 1}^{D^{\ell}} p\left(\bfw^{\ell}_d\right) \underbrace{\Normal{\bfv^{\ell}_d}{\psi(\bfU^{\ell})\bfw^{\ell}_d}{\left[\*\Lambda^{\ell}_d\right]^{-1}}}_{t^{\ell}_d(\bfw^{\ell}_d)}.
\end{equation}
This mirrors the structure of the true posterior in the sense that it is equivalent to the product of the prior and an \textit{approximate likelihood}, $t^{\ell}_d(\bfw^{\ell}_d)$. The variational parameters $\phi$ of this approximation are the parameters of each approximate likelihood, $\bfv^{\ell}_d \in \R^M$ and $\*\Lambda_d^{\ell}\in \R^{M \times M}$---which themselves can be interpreted as \textit{pseudo observations}---and the \textit{global inducing locations}, $\bfU_0 \in \R^{M \times D_0}$, which are used to define $\{\bfU^{\ell}\}_{\ell = 1}^L$ according to
\begin{equation}
    \bfU^1 = \bfU^0 \bfW_1, \quad \bfU^{\ell} = \psi(\bfU^{\ell - 1})\bfW^{\ell} \quad \ell = 2, \ldots, L.
\end{equation}
Optimisation of $\phi$ is achieved through maximisation of the evidence lower bound (ELBO):
\begin{equation}
\label{eq:elbo}
    \mcL_{\text{ELBO}}(\phi; \mcD) = \Exp{q_{\phi}(\bfW)}{\log p(\bfy | \bfW, \bfX)} - \KL{q_{\phi}(\bfW)}{p(\bfW)}.
\end{equation}
We refer to this variational approximation as pseudo-observation variational inference for BNNs (POVI-BNN).

\cite{ober2021global} demonstrate the efficacy of POVI-BNNs relative to mean-field Gaussian variational approximations for BNNs, achieving state-of-the-art performance on a number of regression and classification experiments. Their effectiveness is further demonstrated by \cite{buibiases}, who shows that the estimate of the marginal likelihood provided by the POVI-BNN approximation is close to the true value, indicating the approximation is close to the true posterior.

\section{Amortising Inference in Bayesian Neural Networks}
In this section, we build upon the work of \cite{ober2021global} to develop an effective method of performing amortised inference in BNNs.

Consider the same variational approximation described in \Cref{subsec:gibnn}, except with diagonal precision matrices $\*\Lambda^{\ell}_d$ and inducing locations $\bfU$ replaced by training locations $\bfX$, such that
\begin{equation}
    q\left(\bfW^{\ell} | \{\bfW^{\ell '}\}_{\ell ' = 1}^{\ell - 1}, \mcD\right) \propto \prod_{d = 1}^{D_{\ell}} p\left(\bfw^{\ell}_d\right) \prod_{n=1}^N \underbrace{\Normal{v^{\ell}_{d, n}}{x^{\ell}_{d, n}}{{\sigma_{d, n}^{\ell}}^2}}_{t^{\ell}_{d, n}(\bfw^{\ell}_d)}
\end{equation}
where
\begin{equation}
    \bfx^1_n = \bfW^1 \bfx_n,\quad \bfx^{\ell}_n = \bfW^{\ell}\psi(\bfx^{\ell - 1}_n)\quad \forall \ell = 2, \ldots, L.
\end{equation}
This form of variational approximation enables the use of per-datapoint amortised inference. Specifically, rather than treating the variational parameters $\{\{\{v^{\ell}_{d, n}, \sigma^{\ell}_{d, n}\}_{n=1}^N\}_{d=1}^{D_\ell}\}_{\ell = 1}^L$ as free-form parameters to optimise, we obtain them by passing each datapoint $(\bfx_n, \bfy_n)$ through an inference network at each layer:
\begin{equation}
    \bfv^{\ell}_{n},\ \log \sigma^{\ell}_n = g^{\ell}_{\phi}(\bfx_n, \bfy_n) \quad \forall \ell = 1, \ldots, L.
\end{equation}
The variational parameters now become those of the inference networks, $\{g^{\ell}_{\phi}\}^{L}_{\ell = 1}$. These variational parameters can be shared across datasets, and can be learned using a meta-dataset $\Xi$ with the training objective:
\begin{equation}
    \phi^* = \argmax_{\phi} \mcL(\{\phi^{\ell}\}_{\ell = 1}^L; \Xi); \quad \mcL(\{\phi^{\ell}\}_{\ell = 1}^L; \Xi) = \frac{1}{|\Xi|} \sum_{i=1}^{|\Xi|} \mcL_{\text{ELBO}}(\{\phi^{\ell}\}_{\ell = 1}^L; \mcD_i)
\end{equation}
where $\mcL_{\text{ELBO}}$ is defined in \Cref{eq:elbo}. We refer to our method as amortised pseudo-observation variational inference for BNNs (A-POVI-BNN).

An important limitation of this approach is that performing stochastic optimisation through mini-batching datapoints is not possible, as to compute the $q(\bfW | \mcD)$ we require passing the entire dataset $\mcD$ through the inference network. Nonetheless, this limitation is only significant for large datasets---provided the entire dataset can be passed through the network at once (i.e.\ in the small to medium-sized dataset regime which we consider here) this is not an issue. At test time, we can obtain an approximate posterior $q(\bfW | \mcD_*)$ with a single pass of the dataset through the inference networks. 

\begin{figure}[t]
\floatconts
  {
  fig:regression_experiment
  }
  {
  \caption{Posterior predictive distributions for A-POVI-BNN (top), A-MFVI-BNN (middle), and ConvCNP (bottom) after training on a meta-dataset of size $|\Xi| = 1$ (left) and $|\Xi| = 100$ (right). Data points are shown as black dots, the predictive mean is shown as a dark blue line and the 95\% confidence interval is shown as shaded blue.}
  }
  {%
    \subfigure[A-POVI-BNN ($|\Xi| = 1$)]{
    \label{fig:apovi_bnn_1}%
      \includegraphics[width=0.5\linewidth]{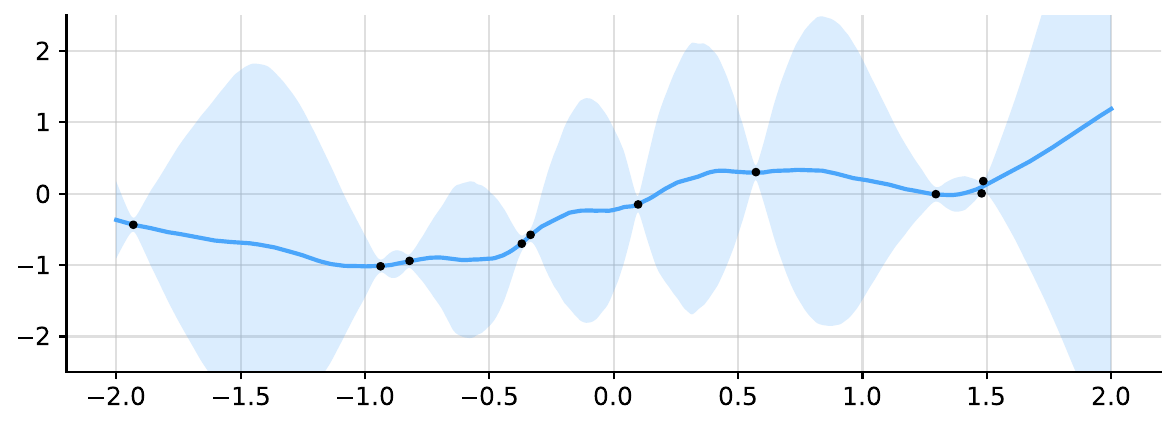}
      }%
    \subfigure[A-POVI-BNN ($|\Xi| = 100$)]{
    \label{fig:apovi_bnn_100}%
      \includegraphics[width=0.5\linewidth]{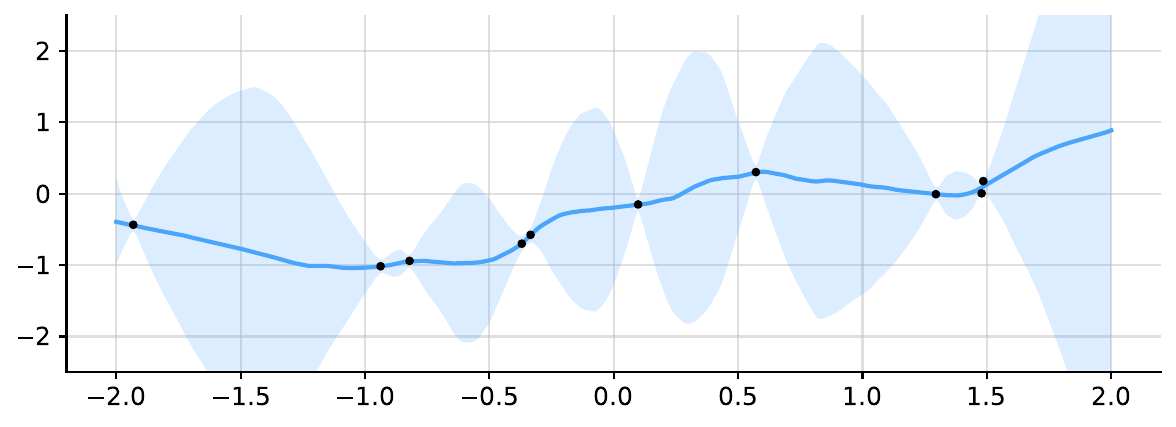}
      }
    \subfigure[A-MFVI-BNN ($|\Xi| = 1$)]{
    \label{fig:amfvi_bnn_1}%
      \includegraphics[width=0.5\linewidth]{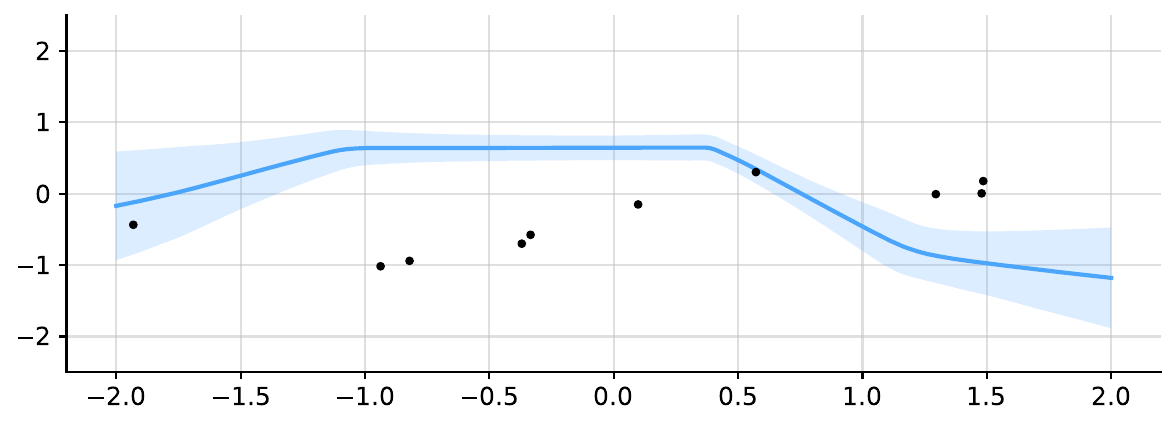}
      }%
    \subfigure[A-MFVI-BNN ($|\Xi| = 100$)]{
    \label{fig:amfvi_bnn_100}%
      \includegraphics[width=0.5\linewidth]{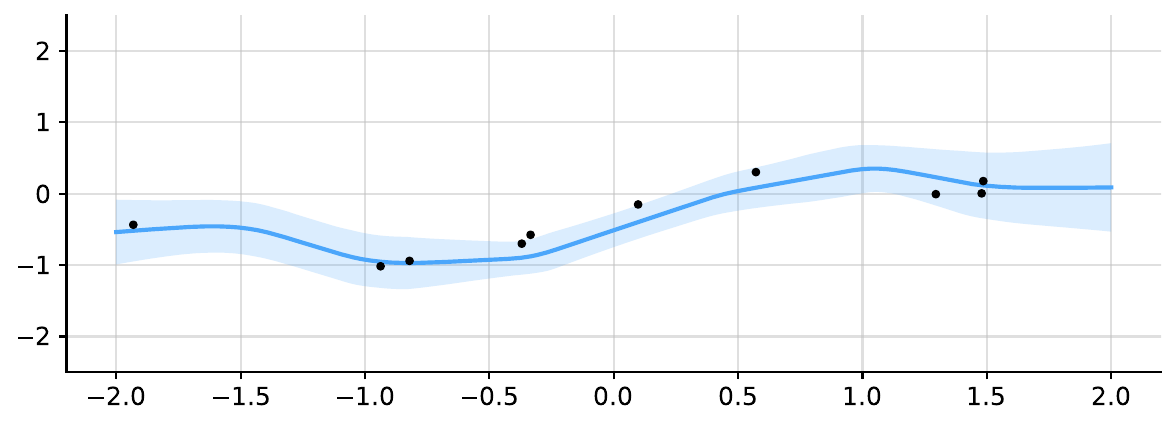}
      }%
      
    \subfigure[ConvCNP ($|\Xi| = 1$)]{
    \label{fig:convcnp_1}%
      \includegraphics[width=0.5\linewidth]{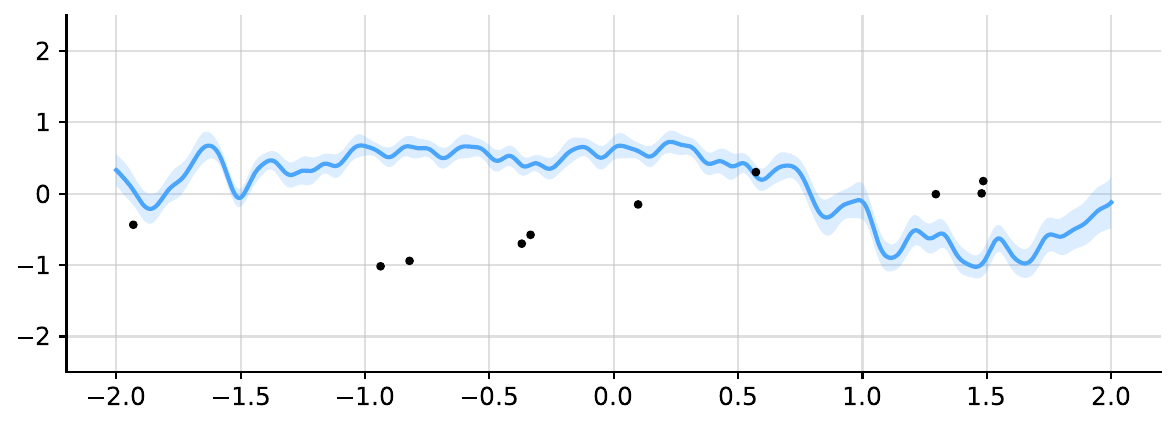}
      }%
    \subfigure[ConvCNP ($|\Xi| = 100$)]{
    \label{fig:convcnp_100}%
      \includegraphics[width=0.5\linewidth]{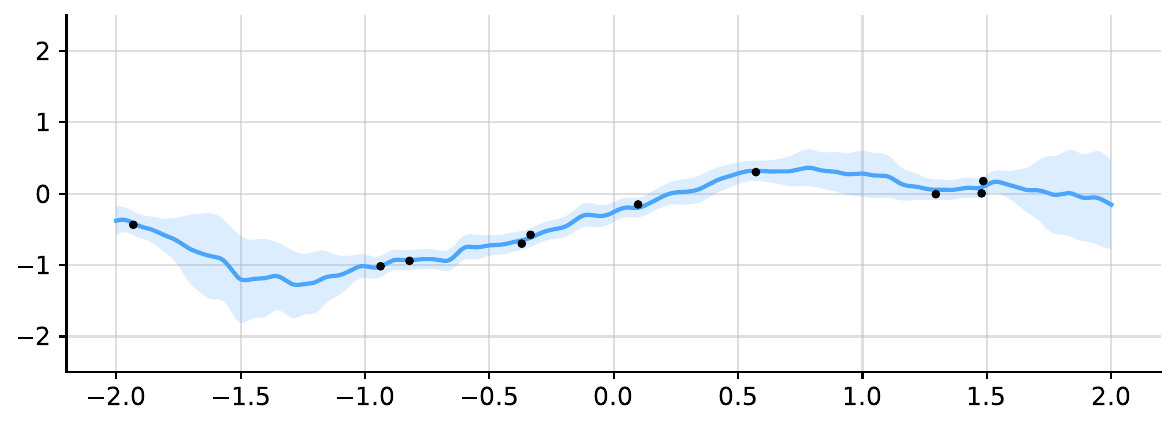}
      }
  }
\end{figure}

\section{Results and Discussion}

In this section, we evaluate the performance of our model in the meta-learning setting. We consider a synthetic meta-dataset consisting of samples from a GP with a squared-exponential (SE) kernel. Each dataset consists of between 10 and 20 training datapoints. We compare A-POVI-BNNs to amortised mean-field variational inference for BNNs (A-MFVI-BNN), which we detail in \Cref{app:amfvi}, and a ConvCNP \citep{gordon2019convolutional}. All NNs used in the amortised BNN architectures (both the model and inference network) consist of two layers of 50 hidden units and ReLU activation functions. The ConvCNP implementation and architecture is identical to that provided in \url{https://github.com/cambridge-mlg/convcnp/blob/master/convcnp_regression.ipynb}.

% \begin{figure}
% \floatconts
%   {
%   fig:apovi_bnn
%   }
%   {
%   \caption{Posterior predictive distributions for the \textbf{A-POVI-BNN}.}
%   }
%   {%
%     \subfigure[$|\Xi| = 1$]{
%     \label{fig:apovi_bnn_1}%
%       \includegraphics[width=0.5\linewidth]{figures/n1/agibnn_se_dataset0_prediction.pdf}
%       }%
%     \subfigure[$|\Xi| = 100$]{
%     \label{fig:apovi_bnn_100}%
%       \includegraphics[width=0.5\linewidth]{figures/n100/agibnn_se_dataset0_prediction.pdf}
%       }
%   }
% \end{figure}

% \begin{figure}
% \floatconts
%   {
%   fig:amfvi_bnn
%   }
%   {
%   \caption{Posterior predictive distributions for the \textbf{A-MFVI-BNN}.}
%   }
%   {%
%     \subfigure[$|\Xi| = 1$]{
%     \label{fig:mfvi_bnn_1}%
%       \includegraphics[width=0.5\linewidth]{figures/n1/amfvi_se_dataset0_prediction.pdf}
%       }%
%     \subfigure[$|\Xi| = 100$]{
%     \label{fig:mfvi_bnn_100}%
%       \includegraphics[width=0.5\linewidth]{figures/n100/amfvi_se_dataset0_prediction.pdf}
%       }
%   }
% \end{figure}

% \begin{figure}
% \floatconts
%   {
%   fig:conv_cnp
%   }
%   {
%   \caption{Posterior predictive distributions for the \textbf{ConvCNP}.}
%   }
%   {%
%     \subfigure[$|\Xi| = 1$]{
%     \label{fig:convcnp_1}%
%       \includegraphics[width=0.5\linewidth]{figures/n1/se_dataset0_prediction.pdf}
%       }%
%     \subfigure[$|\Xi| = 100$]{
%     \label{fig:convcnp_100}%
%       \includegraphics[width=0.5\linewidth]{figures/n100/se_dataset0_prediction.pdf}
%       }
%   }
% \end{figure}

\Crefrange{fig:apovi_bnn_1}{fig:convcnp_100} compare the posterior predictive distributions of each method on an unseen dataset drawn from a GP with the same hyperparameters as those in the meta-dataset. We evaluate the predictions of each method trained on a meta-dataset of size $|\Xi| = 1$ and $|\Xi| = 100$. We see that only A-POVI-BNN obtains a sensible predictive posterior on the unseen dataset in both cases. The ConvCNP performs poorly for $|\Xi| = 1$, which is unsurprising given the large number of model parameters increasing it susceptibility to overfitting. The A-MFVI-BNN performs significantly better for $|\Xi| = 100$, yet, in both cases the quality of its predictive posterior is poor relative to both the A-POVI-BNN and ConvCNP.

Despite these results being very preliminary, they are encouraging and suggest that A-POVI-BNNs may provide a more effective alternative to NPs when the size of meta-datasets are small. We intend to explore the effectiveness of A-POVI-BNNs in more diverse settings, such as image completion, in future work.

% \acks{Acknowledgements go here.}

\bibliography{bibliography}

\appendix

\section{Relationship to Neural Processes}\label{app:neural_processes}

The NP family \citep{garnelo2018conditional,garnelo2018neural} describes the class of meta-learning models which use neural networks to map from datasets to posterior predictive distributions in a single forward pass. Broadly speaking, NPs fall into two categories---conditional NPs (CNPs) and latent NPs (LNPs). In both cases, each observation in the context dataset $\mcD_c$ is processed separately using an \textit{encoder}, $\bfr_c = e_{\theta_e}(\bfx_c, \bfy_c) \ \forall (\bfx_c, \bfy_c) \in \mcD_c$, where $\theta_e$ parameterises a NN. The local encodings $\{\bfr_c\}_{c=1}^C$ are combined into a single representation $\bfr(\mcD_c)$ using a permutation invariant aggregation function, such as summation. This encoding is then passed to the model decoder $d_{\theta_d}$, where $\theta_d$ also parameterises a NN, such that $d_{\theta_d}(\cdot; \bfr(\mcD_c))$ maps from target inputs $\bfX_T$ to predictive distributions. CNPs assume the posterior predictive distribution $p(\bfy_T | \mcD_C, \bfX_T)$ factorises as
\begin{equation}
    p(\bfy_T | \mcD_C, \bfX_T) = \prod_{t =1}^T p_{\theta}(\bfy_{t} | d_{\theta_d}(\bfx_t; \bfr(\mcD_c))).
\end{equation}
LNPs differ to CNPs through their use of a stochastic latent variable $\bfz \sim p_{\theta}(\bfz | \bfr(\mcD_c))$, rather than a fixed embedding:
\begin{equation}
    p_{\theta}(\bfy_T | \bfX_T, \mcD_C) = \int p_{\theta}(\bfz | \bfr_C) \prod_{t \in \mcT} p_{\theta}(\bfy_t | d_{\theta_d}(\bfx_t; \bfz)) d\bfz.
\end{equation}
Importantly, doing so enables dependencies to be maintained across target locations inducing by marginalisation over $\bfz$.

\subsection{Training Neural Processes}
Members of the NP family are trained through minimising the empirical loss of the meta-dataset $\*\Xi$:
\begin{equation}
    \theta^* = \argmin_{\theta} \mcL(\theta; \Xi);\quad \mcL(\theta; \Xi) := \frac{1}{M} \sum_{m=1}^M \mcL(\theta; \mcD_m).
\end{equation}
There are a variety of ways in which the loss function $\mcL(\theta; \mcD_m)$ can be constructed. Focusing on LNPs, the two predominant methods are variational inference (VI) and maximum likelihood (ML). VI training employs the loss function
\begin{equation}
    \mcL(\theta; \mcD) = - \Exp{p_{\theta}(\bfz | \mcD_c \cup \mcD_T)}{\log p_{\theta}(\bfy_T | \bfX_T, \bfz)} + \KL{p_{\theta}(\bfz | \mcD_c \cup \mcD_T)}{p(\bfz | \mcD_c)}
\end{equation}
where $\mcD_c$ and $\mcD_T$ are two random non-overlapping partitions of the dataset $\mcD$. This objective is motivated by the interpretation of $p_{\theta}(\bfz | \mcD)$ as an approximation to the posterior distribution over $\bfz$ given $\mcD$. If $p_{\theta}(\bfz | \mcD)$ is a good approximation to the posterior, then this objective becomes equivalent to the standard VI objective (ELBO):
\begin{equation}
    \mcL(\theta; \mcD) = - \Exp{p_{\theta}(\bfz | \mcD)}{\log p_{\theta}(\bfy_T | \bfX_T, \bfz)} + \KL{p_{\theta}(\bfz | \mcD)}{p(\bfz)}.
\end{equation}
Nonetheless, this objective potentially wastes modelling power on obtaining consistent posterior approximations, rather than good posterior predictive distributions. Instead, \cite{foong2020meta} argue that the following ML objective should be used instead:
\begin{equation}
    \mcL(\theta; \mcD) = - \log p_{\theta}(\bfy_T | \bfX_T, \mcD_c).
\end{equation}
An important difference between the two objectives is that in the VI objective, the parameters of the embedding function can be interpreted as variational parameters, whereas in this ML objective all parameters are model parameters.

\subsection{Amortised Inference in BNNs as Neural Processes}
There are distinct similarities between the model introduced in the previous section and members of the latent neural process (LNP) family. In both cases we 
\begin{enumerate}
    \item compute per-datapoint embeddings of the dataset;
    \item perform a permutation invariant aggregation of embeddings to get a single embedding for the entire dataset;
    \item use this embedding to define a distribution over a latent variable;
    \item sample this latent variable to define mappings from input location to predictive distributions.
\end{enumerate}
The only principal difference between LNPs and our model is the use of shared model parameters in the decoder of LNPs. In our method, the \textit{latent variable is the parameters of the decoder}. Alternative models such as CNAPs \citep{requeima2019fast} and ML-PIP \citep{gordon2018meta} are similar in this manner, except that the per-dataset latent variables are only a small subset of the decoder parameters.

Given these similarities, it is natural to consider training our model in an identical manner to LNPs---in particular, using the ML objective. In doing so, the parameters of the inference network no longer act as variational parameters but rather become part of the model, and so increase the propensity of the model to overfit. We leave exploration of this to future work.

\section{Amortised Mean-Field Variational Inference for BNNs}\label{app:amfvi}
A popular variational approximation for BNNs is the mean-field Gaussian approximation, given by
\begin{equation}
    q(\bfW) = \prod_{\ell = 1}^L \prod_{i = 1}^{|\bfW^{\ell}|} \Normal{w_i}{\mu_i}{\sigma_i^2}
\end{equation}
where $|\bfW^{\ell}|$ denotes the number of elements of weight matrix $\bfW^{\ell}$. We cannot directly apply per-datapoint amortisation to this form of the mean-field approximation as there are not per-datapoint variational parameters. However, we can obtain per-datapoint variational parameters by instead constructing the variational approximation as
\begin{equation}
    q(\bfW) \propto p(\bfW) \prod_{n=1}^N t_n(\bfW)
\end{equation}
where each approximate likelihood $t_n(\bfW)$ is an unnormalised mean-field Gaussian distribution over $\bfW$. Provided $p(\bfW)$ is also mean-field Gaussian (which is typical), then $q(\bfW)$ will be mean-field Gaussian. The variational parameters of each $t_n(\bfW)$ can then be obtained by passing each data pair $(\bfx_n, \bfy_n)$ through an inference network, achieving per-datapoint amortisation.

\end{document}